\documentclass[sigconf,10pt]{acmart}

\usepackage{balance,url,textcomp,mathtools}
\usepackage[linesnumbered,ruled]{algorithm2e}
\usepackage{subcaption}
\usepackage{multirow}
\usepackage{float}
\usepackage{nopageno}

\AtBeginDocument{%
  \providecommand\BibTeX{{%
    \normalfont B\kern-0.5em{\scshape i\kern-0.25em b}\kern-0.8em\TeX}}}

\newcommand{\squishlist}{
 \begin{list}{$\bullet$}
  { \setlength{\itemsep}{0pt}
     \setlength{\parsep}{3pt}
     \setlength{\topsep}{3pt}
     \setlength{\partopsep}{0pt}
     \setlength{\leftmargin}{1.5em}
     \setlength{\labelwidth}{1em}
     \setlength{\labelsep}{0.5em} } }

\newcommand{\squishlisttwo}{
 \begin{list}{$\bullet$}
  { \setlength{\itemsep}{0pt}
     \setlength{\parsep}{0pt}
    \setlength{\topsep}{0pt}
    \setlength{\partopsep}{0pt}
    \setlength{\leftmargin}{2em}
    \setlength{\labelwidth}{1.5em}
    \setlength{\labelsep}{0.5em} } }

\newcommand{\squishend}{
  \end{list}}

\acmYear{2022}\copyrightyear{2022}
\setcopyright{acmlicensed}
\acmConference[SenSys '22]{Workshop on Challenges in Artificial Intelligence and Machine Learning for Internet of Things (AIChallengeIoT)}{November 6--9, 2022}{Boston, MA, USA}
\acmBooktitle{Workshop on Challenges in Artificial Intelligence and Machine Learning for Internet of Things (AIChallengeIoT) (SenSys '22), November 6--9, 2022, Boston, MA, USA}
\acmPrice{15.00}
\acmDOI{10.1145/3560905.3568302}
\acmISBN{978-1-4503-9886-2/22/11}

\begin{document}
\pagenumbering{gobble}
\title{Security Analysis of SplitFed Learning}

\author{Momin Ahmad Khan}
\affiliation{\small{University of Massachusetts, Amherst}}
\email{makhan@umass.edu}

\author{Virat Shejwalkar}
\affiliation{\small{University of Massachusetts, Amherst}}
\email{vshejwalkar@cs.umass.edu}

\author{Amir Houmansadr}
\affiliation{\small{University of Massachusetts, Amherst}}
\email{amir@cs.umass.edu}

\author{Fatima M. Anwar}
\affiliation{\small{University of Massachusetts, Amherst}}
\email{fanwar@umass.edu}

\begin{abstract}
Split Learning (SL) and Federated Learning (FL) are two prominent distributed collaborative learning techniques that maintain data privacy by allowing clients to never share their private data with other clients and servers, and find extensive IoT applications in smart healthcare, smart cities and smart industry.
Prior work has extensively explored the security vulnerabilities of FL in the form of poisoning attacks. To mitigate the effect of these attacks, several defenses have also been proposed.
Recently, a hybrid of both learning techniques has emerged (commonly known as \textbf{\emph{SplitFed}}) that capitalizes on their advantages (fast training) and eliminates their intrinsic disadvantages (centralized model updates).

In this paper, we perform the first empirical analysis of \textbf{SplitFed's} robustness to strong model \emph{poisoning attacks}.
We observe that the model updates in SplitFed have significantly smaller dimensionality
as compared to FL that is known to have curse\ of\ dimensionality. We show that large models that have higher dimensionality are more susceptible to privacy and security attacks, whereas the clients in  SplitFed do not have the complete model and have lower dimensionality, making them more robust to existing model poisoning attacks.
Our results show that the accuracy reduction due to the model poisoning attack is 5x lower for SplitFed compared to FL. 

\end{abstract}

\maketitle
\pagestyle{plain}

\section{Introduction}
\label{sec:introduction}
\textbf{Federated Learning (FL)}~\cite{mcmahan2017communication} is a widely used distributed machine learning setting where several clients collaborate to jointly train a machine learning model by only sharing model updates with a central server.
\textbf{FL} poses two main weaknesses: (1) server has complete access to client model updates and (2) client has to train over the complete model, making it inefficient for resource-constrained devices.
On the other hand another paradigm of distributed machine learning is \textbf{Split Learning (SL)}~\cite{gupta2018distributed} where the model is split between the client and server. This reduces the computational burden on the client device, making it more suitable for resource-constrained devices. Also, the client and the server do not have access to each other's model portions, making \textbf{SL} inherently better in terms of privacy. The disadvantage of \textbf{SL} is that the clients do not perform computation in parallel. One client communicates with the server at any instance, and the rest of the clients have to wait for their turn. 

\textbf{SplitFed}~\cite{thapa2022splitfed} is a new way of distributed collaborative machine learning that merges the above-mentioned systems. It adopts the advantages of both paradigms while minimizing the drawbacks of these paradigms. 
Similar to \textbf{SL}, the model is split between the client and the server, thereby allowing resource-constrained devices to perform computation. An additional server called the \emph{Fed Server} takes the model updates of every client, computed in parallel, at each training round and aggregates them to tune the global client portion of the model, just like in \textbf{FL}. These properties of \textbf{SplitFed} Learning make it suitable for edge deployments.

There is significant research on the security of \textbf{FL} exploring model \emph{poisoning attacks} which perturb the model gradients on malicious devices to reduce accuracy ~\cite{shejwalkar2021manipulating, 247652, shejwalkar2022back, baruch2019little}.
The higher the dimensionality of a model makes the model susceptible to an attack because the adversary has more space to perturb the model in a stealthy manner.
Authors in ~\cite{chang2019cronus} have explored this curse of dimensionality in \textbf{FL} with poisoning attacks.
However, there has been limited research on Split learning security. Authors in ~\cite{pasquini2021unleashing} used  inference attacks ~\cite{hitaj2017deep} 
to recover private training data through either the client or server.
SplitFed has the ability to vary the portion of the model on the client device which may affect the success of existing attacks.
Due to the limitations imposed by dimensionality on the efficacy of model poisoning attacks, an adversary would have a smaller attack space in SplitFed as compared to Federated Learning. 

Based on this relation between \textit{model dimensionality} and \textit{attack success}, we hypothesize that the existing \emph{attacks on the robust aggregation mechanisms would be less effective in a SplitFed}.
In this work, we present the first, detailed security analysis of SplitFed Learning. To test the validity of the our hypothesis 
we conduct an empirical study over a large pool of clients with various combinations of datasets, model architectures, model split layers and attack-defense pairs. 

From our experimental results, we can see that the point at which we split the model has a significant effect on the success of the attack.
Decreasing the client model portion hugely reduces the accuracy drop caused by the attack. We also observe the effect of varying the percentage of malicious clients and see that we start to get a significant accuracy drop at 10\% malicious clients. We evaluate both IID(Independent and Identically Distributed) and non-IID datasets and find that the attacks are stronger on the IID data. We conclude by providing future directions on developing stronger attacks for SplitFed Learning.

Our contributions can be summarized as follows:
\begin{enumerate}
    \item We adapt and launch  the existing state-of-the-art model poisoning attacks in Federated Learning to SplitFed Learning. We implement the optimization-based attack ~\cite{shejwalkar2021manipulating}
    and observe the effects of these attacks when there is (1) no defense mechanism; (2) trimmed mean aggregation rule; and (3) 
    median aggregation rule. 
    \item We explore the effect of the model split location to  answer the following questions: Does varying the layer at which the model is split between the client and the server have an effect on the attack success?
    \item Based on our analysis, we provide recommendations for the possibility of a stronger attack on SplitFed by exploiting its intermediate outputs.
\end{enumerate}

\section{Background}
\label{sec:background}
\subsection{Federated Learning}\label{subsec:federatedlearning}
Federated Learning is a decentralized machine learning mechanism where each client has the same copy of a neural network and its own training data. In each round of training, each client computes forward and backward passes over its model to calculate updated weights. It then sends the updated weights to a central server where all other clients pool their updated weights as well. The central server then performs an aggregation algorithm, such as FedAvg ~\cite{mcmahan2017communication}, and computes a global update. This global update is nothing but the average of all the clients' individual weights. This global update is then shared with each of the clients who start the next round of training with the same weights. This is performed continuously until the model converges. Federated learning thus achieves model training without sharing local training data with other clients or the server involved in the training process.

People have exploited the security vulnerabilities of Federated Learning and shown that it is susceptible to model poisoning attacks that severely degrade the accuracy of the model ~\cite{247652,shejwalkar2021manipulating}.
The LIE(Little is Enough) Attack ~\cite{baruch2019little} takes the mean \(\mu\) and the standard deviation \(\sigma\) of the updates and crafts a malicious update 
\begin{math}
    k = \mu + z*\sigma
\end{math}
where the coefficient \(z\) controls the amount of deviation.
Much stronger than these attacks is the dynamic optimization-based attack proposed by ~\cite{shejwalkar2021manipulating} and we use this attack in our study. This attack is described in detail in Section ~\ref{subsec:optimization attack}.

\subsection{Split Learning}\label{subsec:splitlearning}
\begin{figure*}[!ht]
    \centering
    \includegraphics[width=\textwidth]{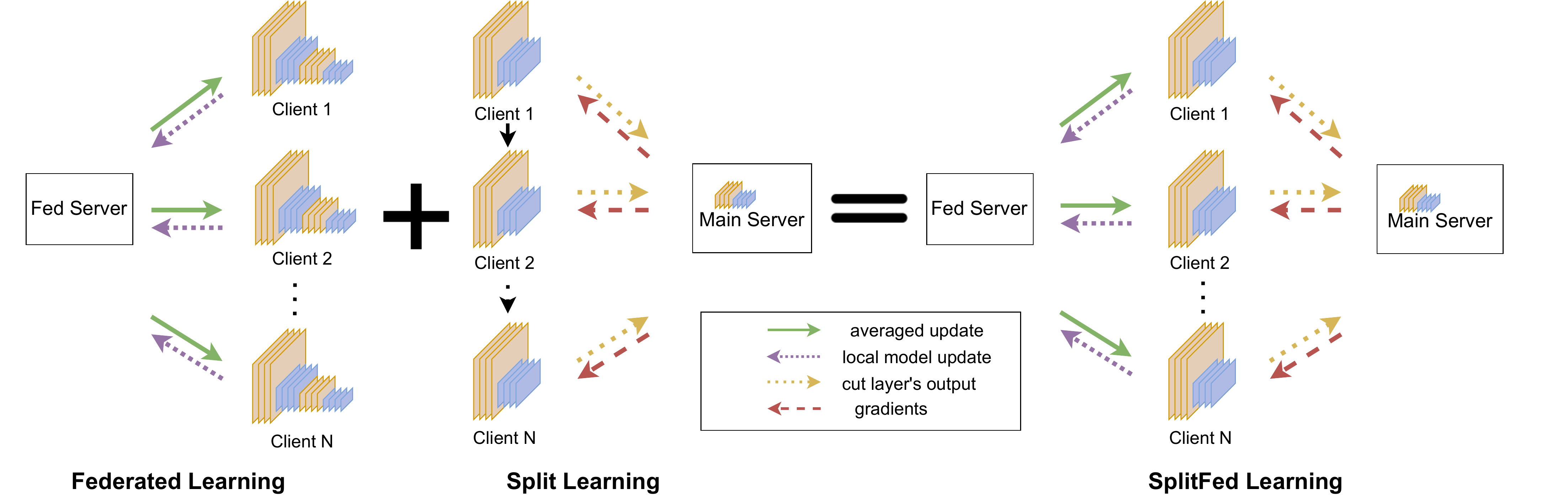}
    \caption{System Model: This figure shows Federated Learning, Split Learning and how they combine to form SplitFed Learning. It should be noted that (1) The model portion on client is different for Split Learning and SplitFed, (2) There is no transfer of model updates between clients in SplitFed.}
    \vspace{-1.5em}
    \label{fig:System_Model}
\end{figure*}
Split Learning ~\cite{gupta2018distributed} is another distributed machine learning framework. It works by splitting the machine learning models between a client and a server. The model is split at some intermediate layer which we will refer to as the \textbf{cut layer} throughout this paper. The output of this cut layer is referred to as \textbf{smashed data}. The initial layers are in the client, while the rest of the layers are on the server.
Therefore, the server performs most of the computation but since it has access to only the second part of the model, it does not have access to the data.

The client performs a forward pass over its portion of the model and sends the smashed data to the server. The server continues the forward propagation over its portion of the model, calculates loss, and begins back-propagation up to the cut layer. It then sends gradients, calculated up to the cut layer, to the client to complete back-propagation. This constitutes one training round of SL. It then sends the updated weights to the next client. The next client starts training from that updated model using its own dataset. The training process is sequential and circular meaning that once we reach the last client, the next client is the first client again.
\subsection{SplitFed Learning}\label{subsection:splitfed}
SplitFed Learning is a new variant ~\cite{thapa2022splitfed} of the distributed machine learning framework which combines both Split Learning and Federated Learning. Although Split Learning enhances model privacy by splitting up the model and the server does not know about the client's side of the model or the data, it is sequential in nature and takes more time to train. Compared to Split Learning, Federated Learning has lower communication efficiency because Split Learning has lesser bandwidth requirements as we are not sharing the complete model at any point.
SplitFed combines the parallel computation from Federated Learning and the resource-efficiency and improved model privacy from Split Learning. The working of the SplitFed model is shown in Figure ~\ref{fig:System_Model}. Similar to the classic Split or Federated Learning settings, we have a set of \(N\) clients and \(N\) models here. Each of the \(N\) models is split at the cut layer between the client and the server. The system model is similar to the Split Learning model except that now we have a FedAvg server on the clients' side, just like in the Federated Learning system model.

Each client computes its forward pass in parallel and sends the intermediate outputs from the \(cut\ layer\) to the server. The server computes the forward and backward passes on each of the clients' outputs and sends them back to the clients. The clients perform the backward pass on their models and send their parameters to the Fed Server. The Fed Server performs Federated Averaging and sends back the parameters to the client.

There is a variant of the SplitFed technique where the server-side computation is done in parallel on each of the client outputs. It completes in lesser time but has reduced accuracy overall. Similar to the client side, it requires an aggregation server on the server side which computes an aggregated server-side model at the end of each global epoch. However, we do not cover that variant in this work.
\vspace{-.8mm}

\section{Poisoning Attacks on SplitFed}
\label{sec:ourapproach}
In this section, we first explain our threat model followed by our adaptations of poisoning attacks and defenses for SplitFed. We formulate optimization based model poisoning attacks under trimmed mean and median model aggregation rule for SplitFed training.

\subsection{Threat Model}
In our threat model, the adversary attempts to compromise model training by injecting malicious updates into the system. The adversary has access to the model on the compromised client. It can modify the weights and gradients of that model. The adversary controls M out of the total N clients. 
In our work, we set M/N to be 0.2 i.e., 20\% of the clients are malicious, which is practical in current deployments.
In addition to this, we assume an honest server. Our assumptions are similar to previous works that have done client-side attacks ~\cite{shejwalkar2021manipulating, pasquini2021unleashing}, where the server is kept honest (from service providers such as Google, Amazon) and provides legitimate service. We also assume that low dimensional model updates are  possible. In a practical setting, this might be the case when multiple hospitals are collaborating to jointly train some cancer detection model and some of those are malicious and wish to corrupt the training process.

The focus of this work is on model poisoning attacks on compromised clients, and  not on data inference attacks on compromised servers~\cite{247652}.
We consider the case where the adversary has knowledge of the model gradients on benign clients, but it can only modify the model gradients on malicious clients. This scenario serves as the upper bound for the efficacy of our attacks and the robustness of our aggregation rules, trimmed mean and median, which we describe in detail in Section ~\ref{subsec:optimization attack}.

\subsection{Optimization-based attack}\label{subsec:optimization attack}
Here we describe the state-of-the-art model poisoning attack which we use in our work. In this type of attack by ~\cite{shejwalkar2021manipulating}, we formulate an optimization problem given by equation \ref{eqn:optimization} that crafts a malicious update which is then sent by the malicious clients to the Fed Server for aggregation. The attacker takes the average of the benign model updates (equation \ref{eqn:fedavg}) and perturbs them with a value \(\gamma*\nabla^p\) as shown in equation \ref{eqn:perturbation} and this is the highest value of \(\gamma\)that bypasses the aggregation rule being used. The optimization problem is formulated as follows:
\begin{equation}
\label{eqn:optimization}
    \underset{\gamma}{\arg\max} = \nabla^{b} - f_{\text{agr}}(\nabla^{m}_{i \in [m]} \cup \nabla_{i \in [m+1, n]})
\end{equation}
Here the average of benign gradients is given by:
\begin{equation}
\label{eqn:fedavg}
    \nabla^b = f_{avg}(\nabla_{{i \in [n]}})
\end{equation}
and each malicious update is given by:
\begin{equation}
\label{eqn:perturbation}
    \nabla^m_{{i \in [m]}} = \nabla^b + \gamma*\nabla^p
\end{equation}
There are different perturbation types, such as unit vector, standard deviation, and opposite direction. In our case, we will be using the standard deviation perturbation as it enables a stronger attack.
\begin{equation}
\label{eqn:std}
    \nabla^p_{std} = -std(\nabla_{{i \in [n]}})
\end{equation}

This attack is designed to break through state-of-the-art robust aggregation mechanisms such as trimmed mean and median. Next, we describe those aggregation mechanisms briefly and how the optimization framework is tailored to break through trimmed mean and median.

\textbf{Trimmed Mean} is an aggregation rule
~\cite{yin2018byzantine}, ~\cite{xie2018generalized} that replaces the simple averaging technique. It takes the set of all updates, including malicious ones, as input, sorts the updates along each dimension \(i\), and removes  $m$ of the largest and smallest values from the set. Note that, in SplitFed, these updates are from the portion of the model on the clients only. Here, $m$ is the number of malicious clients. Then it takes the mean of the updates along each trimmed dimension. The aggregated update is then sent back to each client.
The general optimization framework given in equation ~\ref{eqn:optimization} is modified for trimmed mean where \(f_{agr}\) is now \(f_{trmean}\) as done by ~\cite{shejwalkar2021manipulating}. Here, the trimmed mean is computed on the aggregate of malicious and benign updates, and the adversary's objective is to maximize the norm between this trimmed mean and the mean of benign updates.

\textbf{Median} is similar to trimmed mean ~\cite{yin2018byzantine}, ~\cite{xie2018generalized} in the sense that it also removes malicious update components along each dimension of the update by means of a statistical operation. It computes the element-wise median of the updates along each dimension. The general optimization framework given in equation \ref{eqn:optimization} is modified in a similar fashion to the trimmed mean attack. The only difference it has is that \(f_{agr}\) is now \(f_{median}\).

\section{Implementation}
\label{sec:implementation}
We implemented our adapted  poisoning attacks for  SplitFed on a variety of image datasets and models, and evaluated over hundreds of clients.
 
\subsection{Datasets}
  
 \noindent \textbf{CIFAR10 ~\cite{krizhevsky2009learning}}
    consists of 60000 images of size 32x32 colored images. There are 10 classes of common things like airplanes, birds, cats, dogs, etc. Among these 60000 images, 50000 are training images while the other 10000 are test images. We chose this dataset because it is commonly used in such works and would serve as a benchmark for easier comparison.
    
    \noindent\textbf{FEMNIST~\cite{caldas2018leaf}}
    consists of 805263 grayscale images distributed over 3550 users. We can partition them in an IID or non-IID fashion. Since we are using CIFAR10 in an IID fashion, we split up FEMNIST among 3500 users in a non-IID fashion. Each grayscale image is 28 x 28 pixels and is one of 62 classes; 26 classes each for upper and lower case letters and 10 classes for digits. The mean number of samples per user is 226.83 and the standard deviation is 88.94. We test FEMNIST in a cross-device setting, where we are choosing 100 random clients out of 3550 every round for training in that particular round.

\subsection{Models}

    \textbf{Alexnet ~\cite{krizhevsky2012imagenet}}
    is a standard model with some convolutional layers followed by a fully connected layer. We train it on the CIFAR10 dataset for 1000 epochs with a learning rate of 0.01, batch size of 64, and an SGD optimizer. We split the model at three different layers. More specifically we split it at the first, second, and third maxpool layers. We call these split versions V1, V2, and V3 respectively.
    
    \noindent \textbf{VGG11 ~\cite{simonyan2014very}} has several convolutional layers followed by a fully connected layer. We train it on the CIFAR10 dataset for 200 epochs with a learning rate of 0.01, batch size of 64, and an SGD optimizer. We split the model at three different layers. More specifically we split it at the second, third, and fourth maxpool layers. We call these split versions V1, V2, and V3 respectively.
    
    \noindent We design a \textbf{Custom}
    CNN for our FEMNIST dataset. The model is composed of two convolutional layers followed by two fully connected layers. The model for FEMNIST is small and simple because the dataset consists of 28x28 grayscale images and we do not need a very large model to fit on this dataset. We use the image dataset for a particular client as a complete batch since it is small in size. This is a small model so the model split at layers 1,2,3 is called V1,V2, and V3 respectively.

\subsection{Reproducibility}
To ensure the reproducibility of our results we fix the random seeds to be 42 for PyTorch, NumPy, and Python's random package. This is because there are several parts of the code where randomness is taking place, such as the dataloader, weight initializations etc. Reproducibility is even more important in our non-IID case of the FEMNIST dataset where we want the same client selection sequence for each experiment trial we conduct.

\subsection{Metrics}
We report the efficacy of the attack in terms of accuracy drop ($Acc_{drop}$) that quantifies the difference between the test accuracy obtained without attack ( \(Acc\))and the test accuracy obtained with attack (\(Acc\textsubscript{$attack$}\))
\begin{equation}\label{eqn:accuracy}
    Acc\textsubscript{$drop$} = Acc - Acc\textsubscript{$attack$}    
\end{equation}
Simply reporting the accuracy after the attack would not 
be a fair comparison because we get different accuracies for different models and model splits.

\subsection{Experimental Setup}
To test the case of a practical setting, we evaluate our setup with 100 clients. With our setting of 20\% malicious clients, 20 of the clients from among our 100 are malicious. We use several NVIDIA RTX8000 GPUS, each having 48GB memory on our university-provided GPU cluster to run our experiments. All the code was written in PyTorch(1.12.1+cu116) and run on Jupyter Notebooks on the GPU cluster. We perform hyperparameter tuning with lr=[0.01, 0.05, 0.1], momentum = [0.9, 0.95, 0.99], and batch\_size = [32, 64, 128].

\section{Analysis}
\label{sec:evaluation}
In this section we will discuss the effect of poison attack in the SplitFed setting on two representative datasets, three models and model splits and two attack-defense pairs.
\subsection{Cut Layer and Attack effectiveness}
We have seen that the choice of different cut layers does not functionally change the split learning system. However, the choice of cut layer plays a huge role in the effectiveness of our attack. We see that when the client has a smaller portion of the full model the attack does not reduce the overall accuracy much. Conversely, when the client has a larger portion of the model, the adversary has a lot of space to manipulate the model. In this scenario, the overall accuracy of the final model is severely degraded. 

\begin{figure}[h]
    \centering
    \begin{subfigure}[b]{0.48\columnwidth}
        \includegraphics[width=\columnwidth]{{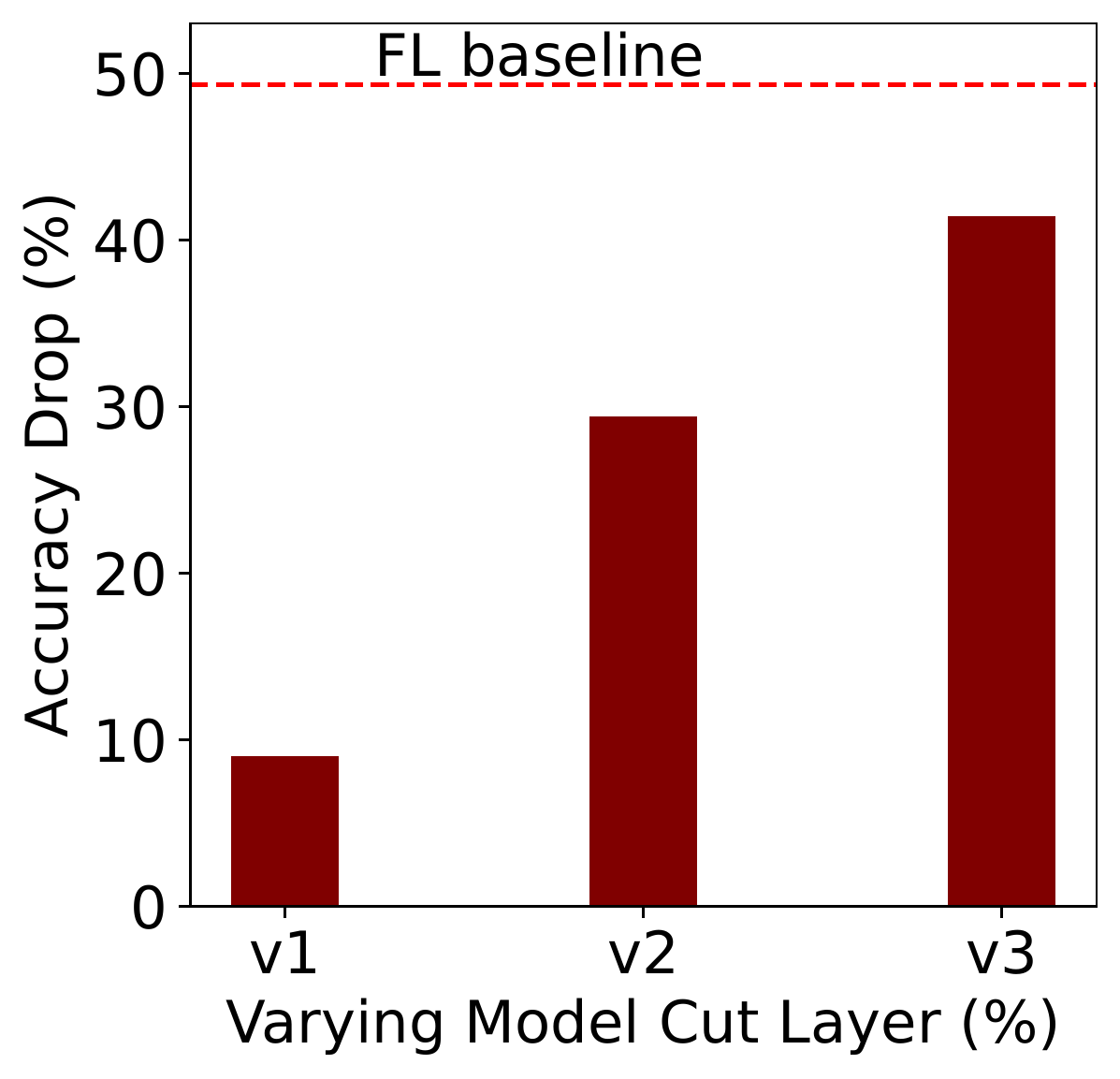}}
        \caption{Alexnet}
        \label{fig:layers_vs_accuracy_alexnet_trmean}
    \end{subfigure}
    \hspace{0.5mm}
    \begin{subfigure}[b]{0.48\columnwidth}
        \includegraphics[width=\columnwidth]{{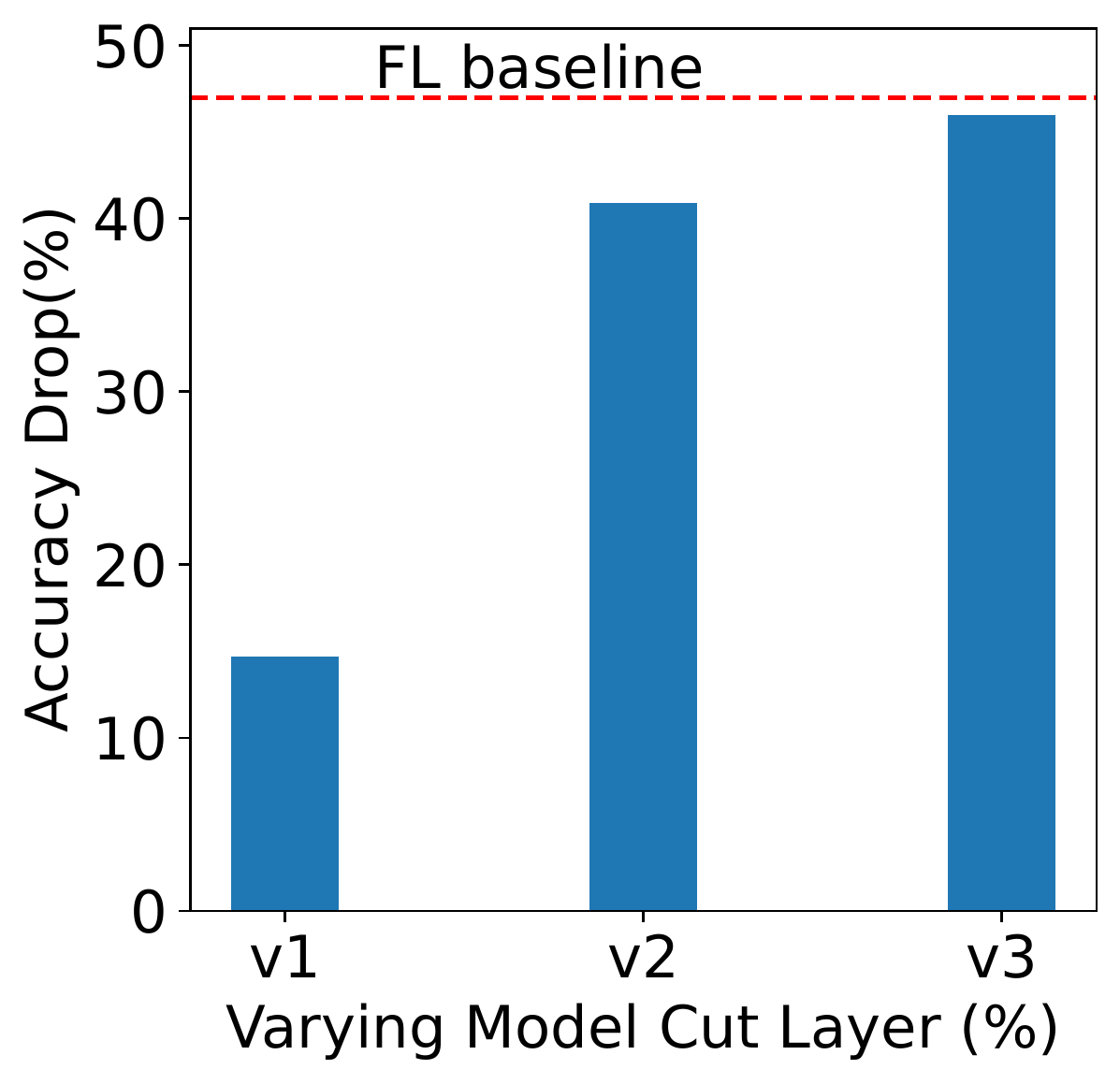}}
        \caption{VGG}
        \label{fig:layers_vs_accuracy_vgg_trmean}
    \end{subfigure}
    \caption{Impact of attacks tailored for trimmed mean defense with varying cut layers on Alexnet and VGG models.}
    \vspace{-6mm}
    \label{fig:layers_vs_accuracy_trmean}
\end{figure}

\begin{figure}[h]
    \centering
    \begin{subfigure}[b]{0.48\columnwidth}
        \includegraphics[width=\columnwidth]{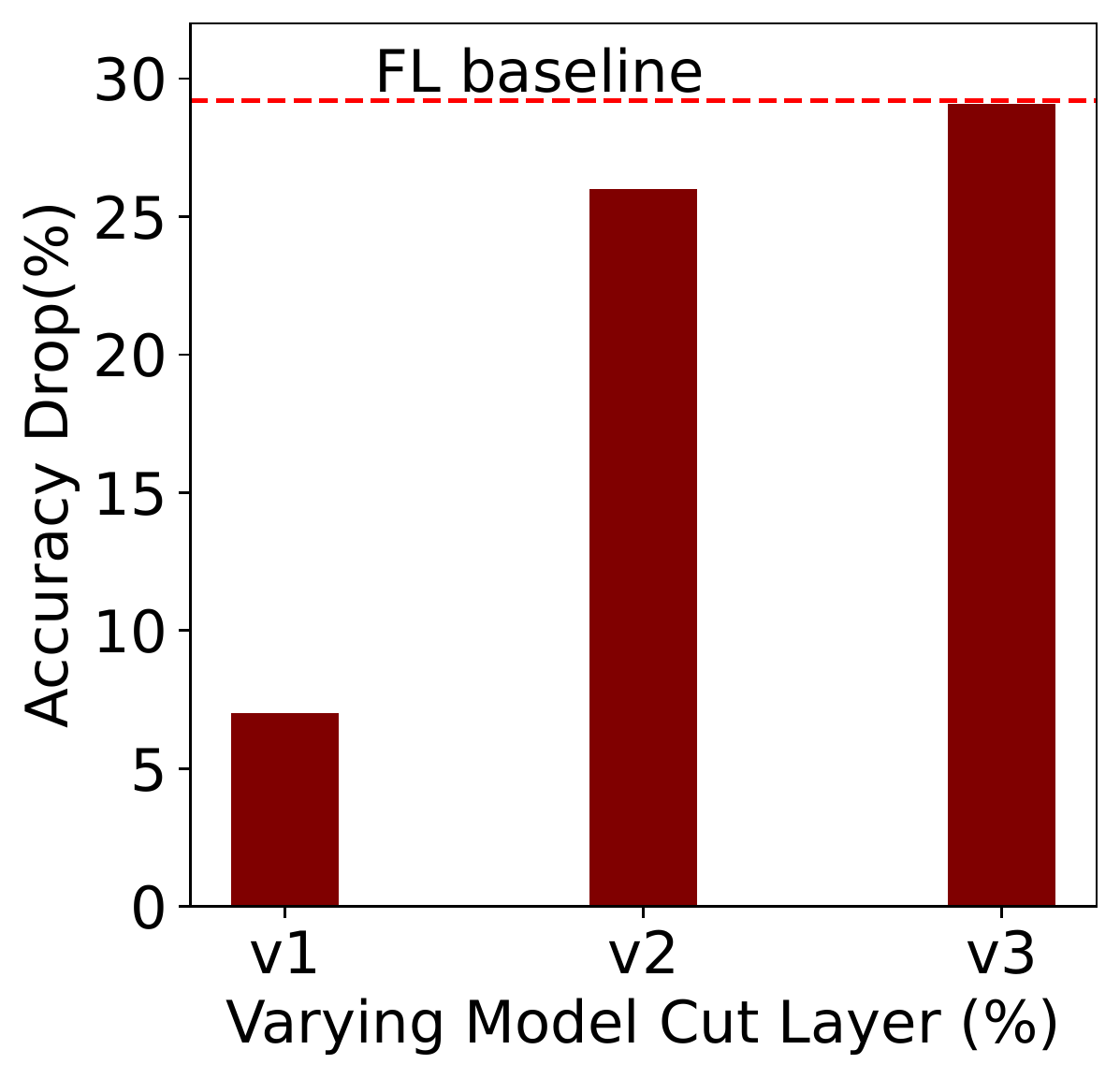}
        \caption{Alexnet}
        \label{fig:layers_vs_accuracy_alexnet_median}
    \end{subfigure}
    \hspace{0.5mm}
    \begin{subfigure}[b]{0.48\columnwidth}
        \includegraphics[width=\columnwidth]{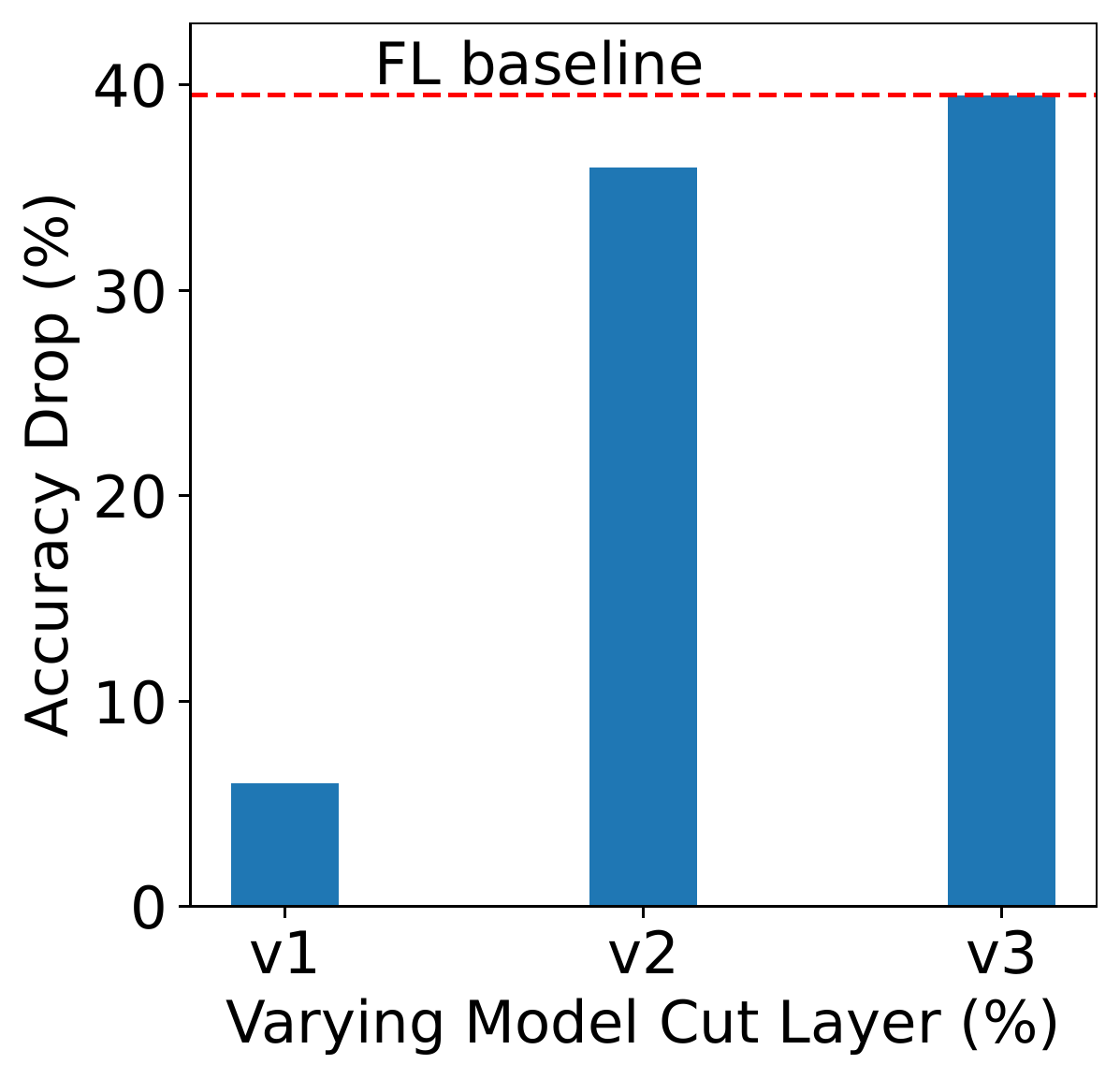}
        \caption{VGG}
        \label{fig:layers_vs_accuracy_vgg_median}
    \end{subfigure}
    \caption{Impact of attacks tailored for median defense with varying cut layers on Alexnet and VGG models.}
    \vspace{-6mm}
    \label{fig:layers_vs_accuracy_median}
\end{figure}

As shown in Table~\ref{tab:splitfed_attacks}, in the case of Alexnet trained in CIFAR10, when we increase the percentage of the model on the client side from VGG v1 to VGG v3, and the accuracy drop increases from \(9.05\%\) to \(41.42\%\) for trimmed mean. When we compare it with the federated learning case in the same setting and it gives a higher accuracy degradation of (\(49.3\%\)) as shown in Table~\ref{tab:fedavg}. because it has the complete model on the client side. Figures ~\ref{fig:layers_vs_accuracy_trmean} and ~\ref{fig:layers_vs_accuracy_median} show this accuracy as the \(FL\ baseline\) (shown as red dashed line). We also observe that the \(v3\) model split has almost equal accuracy drop as the baseline (FL) however the accuracy drop of \(v2\) split is lower than \(v3\) but still significantly higher than that of \(v1\).
Based on these results, we can advocate the use of \(v1\) versions of our models for more robustness against poisoning attacks. This can also prove to be beneficial for resource-constrained devices and could lead to faster training time.

\begin{table}[t]
\begin{tabular}{|l|ll|ll|}
\hline
\multirow{2}{*}{\begin{tabular}[c]{@{}l@{}}Dataset\\ Model\end{tabular}} & \multicolumn{2}{l|}{TrMean} & \multicolumn{2}{l|}{Median} \\ \cline{2-5} 
 & \multicolumn{1}{l|}{Acc} & Acc\textsubscript{$drop$} & \multicolumn{1}{l|}{Acc} & Acc\textsubscript{$drop$} \\ \hline
\begin{tabular}[c]{@{}l@{}}CIFAR10\\ VGG\end{tabular} & \multicolumn{1}{l|}{73} & 47 & \multicolumn{1}{l|}{71.5} & 39.5 \\ \hline
\begin{tabular}[c]{@{}l@{}}CIFAR10\\ Alexnet\end{tabular} & \multicolumn{1}{l|}{62.4} & 49.3 & \multicolumn{1}{l|}{62.1} & 29.19 \\ \hline
\begin{tabular}[c]{@{}l@{}}FEMNIST\\ Custom\end{tabular} & \multicolumn{1}{l|}{81} & 18 & \multicolumn{1}{l|}{75} & 8 \\ \hline
\end{tabular}
\vspace{-.5mm}
\caption{Attacks in the Federated Learning Setting}
\vspace{-10mm}
\label{tab:fedavg}
\end{table}
\subsection{Impact of Defense}
Next, we compare the two defenses, the trimmed mean and median in our evaluation as shown in Tables~\ref{tab:fedavg} and Table ~\ref{tab:splitfed_attacks}. We can see throughout all settings with different model architectures, datasets, and model splits median defense performs better than the trimmed mean defense. The highest performance difference is for Alexnet (v3), where the median is about \(11\%\) better than trimmed mean. For the Federated case, we see the same trend with median with \(8-10\%\) better performance than the trimmed mean. This trend is consistent with FEMNIST too for both FL and SplitFed. 
From this we can conclude that the median defense is more robust than the trimmed mean defense. This may be due to reason that median is a much more robust measure of central tendency than mean in the presence of an outlier that can still exist when the upper and lower m values are trimmed. Therefore, it is much harder for the adversary to craft an update that bypasses the median defense.

\subsection{IID vs non-IID}
Initially we perform the attack on non-IID FEMNIST with the same learning rate as in IID CIFAR10(lr=0.01). Similarly to our IID case, we do poisoning at every training round. This leads to very low model accuracy (below $10\%$/) because the adversary gets too much attack space in non-IID because the model updates are very different from each other, and the high learning rate allows it to craft a malicious update of large magnitude. This effect is also validated in ~\cite{shejwalkar2021manipulating, 247652}. We then perform the attack with a lower learning rate of $0.001$ and carry out poisoning after several training rounds without poisoning. We report these results in Tables~\ref{tab:fedavg} and Table~\ref{tab:splitfed_attacks}. A significant observation in this case is when we use the $v1$ model split, where we do not get a drop in accuracy with our attack. This is because the model we are using is already small in size, and the adversary gets an even smaller part to attack.

\subsection{Impact of the Percentage of Malicious Clients}
\begin{figure}[!htb]
    \centering
    \includegraphics[width=.8\columnwidth]{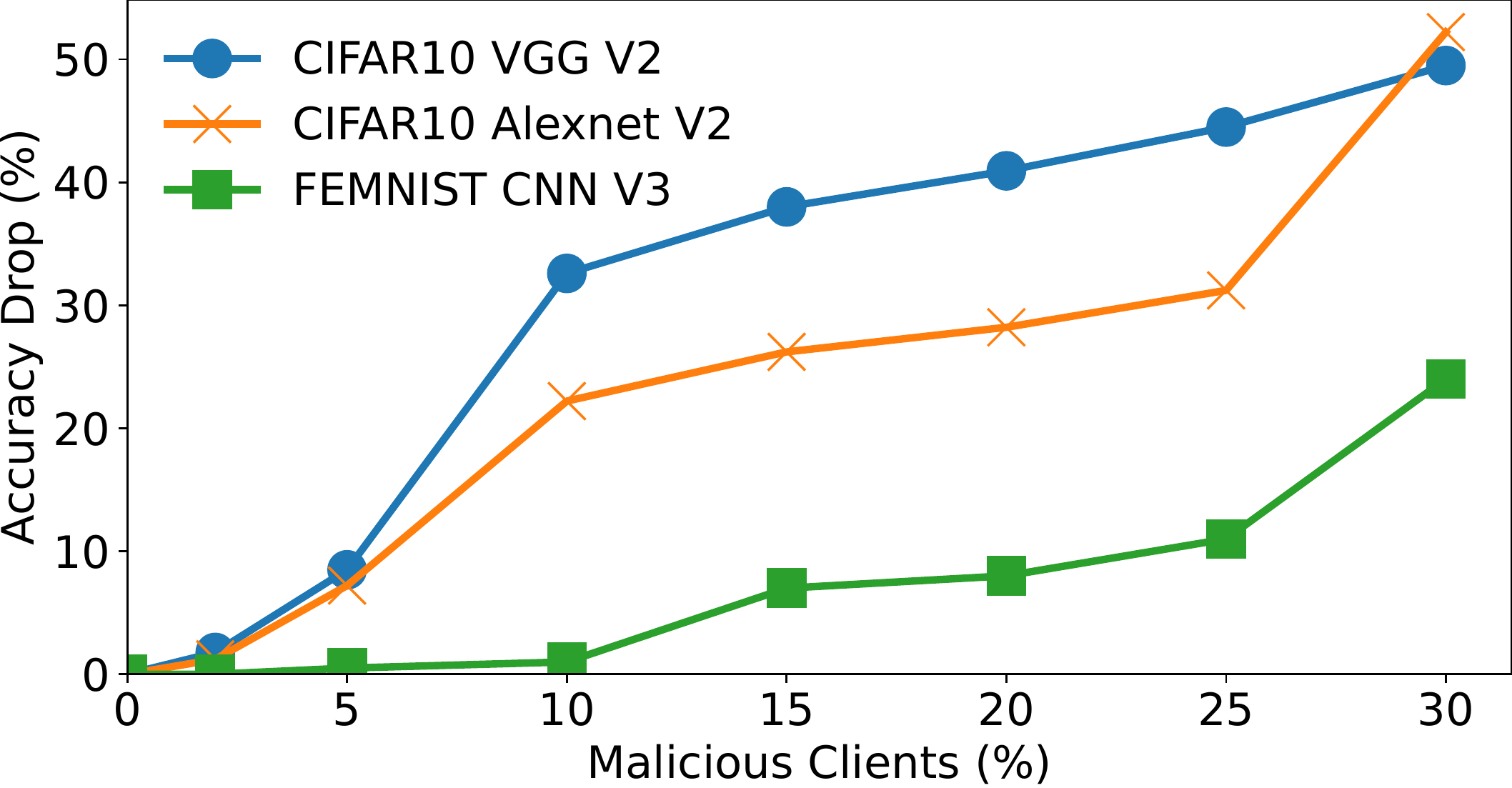}
    \caption{Effect of varying percentage of malicious clients on accuracy drop.}
    \vspace{-6mm}
    \label{fig:percentage_malicious_clients}
\end{figure}

We need to determine what percentage of clients would be suitable to perform significant accuracy degradation. Although it is intuitive that to get the highest degradation, we need a higher percentage of malicious clients, we need to keep practical considerations in mind. It is not possible to have a very large number of malicious clients in a practical setting ~\cite{shejwalkar2022back}. We tested our attack with a varying percentage of malicious clients for the CIFAR10 dataset. We see that the accuracy of the model starts to degrade with a tiny percentage of malicious clients (2\%) and that the accuracy of the model shows a sharp degradation at 10\% malicious clients. For the Alexnet V2 model, we see a complete degradation in model accuracy when we increase the percentage of malicious clients to 30\%. In this scenario, the accuracy of the model is about 10\% which is equivalent to a random guess for a classification problem of 10 classes, these trends are comparable with Federated Learning as well, as reported in ~\cite{shejwalkar2021manipulating}. 
We observe that the same increase in the percentage of malicious clients in Federated Learning causes a much larger change in accuracy drop as compared to SplitFed. Increasing the percentage of malicious clients from $10\%$ to $15\%$ for CIFAR10 with Alexnet and trimmed mean increases the accuracy drop by $11\%$ in Federated Learning while the accuracy drop in SplitFed is $4\%$, as shown in Figure~\ref{fig:percentage_malicious_clients}. We see this effect for every $5\%$ increment in malicious clients from $0\%$ to $30\%$. This observation is also consistent with the FEMNIST dataset.

\begin{table}[t]
\begin{tabular}{|c|c|cc|cc|}
\hline
\multirow{2}{*}{\begin{tabular}[c]{@{}c@{}}Dataset\\ Model\end{tabular}} &
  \multicolumn{1}{l|}{\multirow{2}{*}{Split}} &
  \multicolumn{2}{c|}{TrMean} &
  \multicolumn{2}{c|}{Median} \\ \cline{3-6} 
 &
  \multicolumn{1}{l|}{} &
  \multicolumn{1}{c|}{Acc} &
  Acc\textsubscript{$drop$} &
  \multicolumn{1}{c|}{Acc} &
  Acc\textsubscript{$drop$} \\ \hline
\multirow{3}{*}{\begin{tabular}[c]{@{}c@{}}CIFAR10\\ VGG\end{tabular}} &
  v1 &
  \multicolumn{1}{c|}{75.68} &
  14.68 &
  \multicolumn{1}{c|}{75.41} &
  6.81 \\ \cline{2-6} 
 &
  v2 &
  \multicolumn{1}{c|}{71.49} &
  40.93 &
  \multicolumn{1}{c|}{71.92} &
  36.92 \\ \cline{2-6} 
 &
  v3 &
  \multicolumn{1}{c|}{67} &
  46 &
  \multicolumn{1}{c|}{65.5} &
  39.5 \\ \hline
\multirow{3}{*}{\begin{tabular}[c]{@{}c@{}}CIFAR10\\ Alexnet\end{tabular}} &
  v1 &
  \multicolumn{1}{c|}{63.05} &
  9.05 &
  \multicolumn{1}{c|}{63.42} &
  7.42 \\ \cline{2-6} 
 &
  v2 &
  \multicolumn{1}{c|}{63.39} &
  29.39 &
  \multicolumn{1}{c|}{62.34} &
  26.34 \\ \cline{2-6} 
 &
  v3 &
  \multicolumn{1}{c|}{61.71} &
  41.42 &
  \multicolumn{1}{c|}{62.1} &
  29.1 \\ \hline
\multirow{3}{*}{\begin{tabular}[c]{@{}c@{}}FEMNIST\\ Custom\end{tabular}} &
  v1 &
  \multicolumn{1}{c|}{87} &
  0 &
  \multicolumn{1}{c|}{87} &
  0 \\ \cline{2-6} 
 &
  v2 &
  \multicolumn{1}{c|}{87.3} &
  2.3 &
  \multicolumn{1}{c|}{87.3} &
  1.6 \\ \cline{2-6} 
 &
  v3 &
  \multicolumn{1}{c|}{84} &
  8 &
  \multicolumn{1}{c|}{84.9} &
  5.2 \\ \hline
\end{tabular}
\caption{Attacks in the SplitFed setting.}
\vspace{-12mm}
\label{tab:splitfed_attacks}
\end{table}
\vspace{-4mm}

\vspace{-4mm}

\section{Discussion and Conclusion}
\label{sec:conclusion}
\subsection{Additional Attack Space}
So far we have performed the attack at the client-side model aggregator, to perform a comparison with Federated Learning. It would be interesting to see the effect of performing some perturbation on the \textit{smashed data} sent by some malicious client to the server. The dimensionality of smashed data would change with different choices of the \textit{cut layer} and this would most certainly affect the success of the attack.

\subsection{Conclusion}
We performed an extensive analysis of model poisoning on the SplitFed system by using existing state-of-the-art attacks. We test these attacks by pairing them with various defenses, models, datasets, data distributions, and model splits and find that the same attacks in SplitFed have lower efficacy than in Federated Learning because in SplitFed the client does not have the complete model. We also propose an additional attack space that can be explored in future work.  
  
\bibliographystyle{ACM-Reference-Format}
\bibliography{paper}

\end{document}